\documentclass[10pt,twocolumn,letterpaper]{article}

\usepackage[pagenumbers]{cvpr}

\usepackage{booktabs}
\usepackage{array}
\usepackage{adjustbox}     %
\usepackage{multirow}
\newcommand{\Ro}{\mathcal{R}}                      %
\newcommand{\Lgeo}{\mathcal{L}_{\mathrm{pos}}}     %
\newcommand{\Ldir}{\mathcal{L}_{\mathrm{dir}}}     %
\newcommand{\Lgrs}{\mathcal{L}_{\mathrm{GRS}}}     %
\newcommand{\Ldist}{\mathcal{L}_{\mathrm{dist}}}   %
\newcommand{\std}[1]{{\fontsize{6}{7}\selectfont$\pm$#1}}   %

\definecolor{cvprblue}{rgb}{0.21,0.49,0.74}
\usepackage[pagebackref,breaklinks,colorlinks,allcolors=cvprblue]{hyperref}

\title{GeoRead3D: Grounding Part Descriptions for 3D Pointing and Reorientation}

\author{
Zhangcheng Hou\\
School of Science and Technology\\
Keio University\\
Yokohama 223-8522, Japan\\
{\tt\small zhangcheng@ohtsuki.ics.keio.ac.jp}
\and
Kevin Tonkich\\
School of Science and Technology\\
Keio University\\
Yokohama 223-8522, Japan\\
{\tt\small kevin@ohtsuki.ics.keio.ac.jp}
\and
Tomoaki Ohtsuki\\
School of Science and Technology\\
Keio University\\
Yokohama 223-8522, Japan\\
{\tt\small ohtsuki@ics.keio.ac.jp}
}

\begin{document}
\twocolumn[{%
\renewcommand\twocolumn[1][]{#1}%
\maketitle
\begin{center}
  \includegraphics[width=\textwidth]{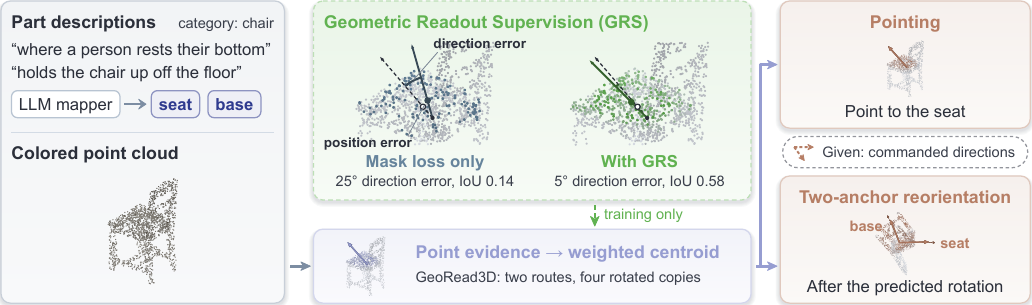}
  \captionof{figure}{\textbf{Learning part evidence for pointing and reorientation with GRS.}
  The weighted centroid of the point evidence, relative to the object center, gives each part's direction for both tasks.
  During training, GRS supervises the centroid's position and direction.
  Matched heads on frozen Utonia features, trained with the mask loss alone or with GRS: with GRS, the evidence covers the part more accurately and its centroid points closer to the true direction (input pose).
  Dashed arrows show true or commanded directions and solid arrows predicted ones.
  White circles mark true part centers and dark dots the weighted centroids.}
  \label{fig:teaser}
\end{center}
}]
\begin{abstract}
Language often refers to an object by its parts, as in asking to point to a part or to turn an arbitrarily posed object so that two named parts lie in given directions.
Both tasks need the direction from the object's center to the part's center, not the part's own orientation.
A natural approach segments the part and takes the centroid of its mask.
However, high mask overlap predicts the pointing direction poorly when a part is spread out relative to its distance from the object center.
Language-guided part segmentation is trained and evaluated by mask overlap, which leaves a gap between the masks it produces and the part geometry that these tasks use.
We close this gap by learning point evidence, weights over the observed points whose weighted centroid gives the part's location.
Geometric Readout Supervision (GRS) trains these weights through the centroid's position and adds direction supervision on the parts used in reorientation commands.
GRS improves both tasks even with a single frozen encoder and a single input pose.
We implement this supervision in GeoRead3D, a system for pointing and reorientation from part descriptions.
A language model maps each description to a known part name.
Dual-Route Part Evidence combines refined segmentation evidence with directly predicted point evidence to locate each part.
On PartNeXt and Artiverse, GeoRead3D points within $22.5^\circ$ for $74.1\%$ and $60.0\%$ of queries, $5.8$ and $7.3$ percentage points above the strongest baselines.
Its reorientation success exceeds the strongest baselines by $10.4$ and $7.8$ points.
\end{abstract}

\section{Introduction}
\label{sec:intro}

Consider a request to point to the part of a chair ``where a person rests their bottom'', or to turn the chair so that its seat and base lie in two given directions.
Either request requires turning a description of a part into a location on an object seen in an arbitrary pose (\cref{fig:teaser}).
Both tasks use the offset from the object's center to the part's center.
Pointing uses its direction, and two-anchor reorientation aligns the directions of two parts, the anchors, with two commanded ones.

Part segmenters that accept text queries~\cite{11444019,Hadgi_2026_CVPR,jin2026cosmo3d,10943567} label the points that match the query.
Multimodal language models~\cite{xu2024pointllm,zhu2026partllmunifiedmultimodalfoundation} describe objects or segment their parts.
A part mask identifies which points belong to a part, while both tasks use a direction derived from the part's location.
A few misplaced points can shift the centroid even when mask overlap is high.
This motivates learning part evidence whose readout supports accurate pointing and reorientation.

We introduce Geometric Readout Supervision (GRS), which learns query-conditioned point evidence through its object-relative position and direction (\cref{fig:teaser}).
These are obtained from the evidence-weighted centroid relative to the object center.
Supervising only the position of this readout couples its length and direction.
The loss then becomes less sensitive to the direction as the predicted readout shrinks, although both tasks depend only on directions.
GRS adds explicit direction supervision and applies it to parts that serve as anchors in reorientation commands.
Controlled experiments show that GRS improves pointing and reorientation even with a single frozen encoder and a single input pose.

A fixed language model maps each description to a part name of a predefined taxonomy, which conditions the part evidence, while GRS addresses the geometric localization of the named part.
We develop Dual-Route Part Evidence.
The refinement route starts from a part segmentation and corrects its evidence for localization.
The direct route provides a second source of point evidence, predicted independently of the segmentation outputs.

Dual-Route Part Evidence assigns the evidence of both routes to the same input points.
This allows GeoRead3D to fuse it across routes and rotated inputs and to apply a shared geometric readout to each queried part.
GeoRead3D leads every compared method on both test sets.
It points correctly for $74.1\%$ of queries on our held out PartNeXt test split and $60.0\%$ on Artiverse~\cite{Iliash_2026_CVPR}, $5.8$ and $7.3$ points above the strongest baseline.
Its lead in reorientation is $10.4$ and $7.8$ points.

\noindent In summary, we make three contributions.
\begin{itemize}[leftmargin=*,itemsep=1pt,topsep=2pt]
  \item \textbf{Geometric Readout Supervision.} We supervise query-conditioned point evidence through the object-relative position and direction of its readout. Direction supervision focuses on parts that can anchor a reorientation, aligning both the training target and the choice of supervised parts with the two tasks.
  \item \textbf{Dual-Route Part Evidence.} We construct complementary part evidence over the same input points, by refining a segmentation and by predicting evidence directly.
  \item \textbf{GeoRead3D.} We connect part descriptions to pointing and reorientation through shared point evidence. Evidence is fused across routes and poses before a single geometric readout for each queried part.
\end{itemize}

\section{Related Work}
\label{sec:related}

\paragraph{Language-guided 3D part segmentation.}
Early methods lift predictions or features of 2D foundation models from rendered views to 3D parts~\cite{10203074,10657376,10943567}.
Find3D~\cite{11444019} and PatchAlign3D~\cite{Hadgi_2026_CVPR} align point or patch features with text.
CoSMo3D~\cite{jin2026cosmo3d} adds supervision in a canonical space for consistency across poses and categories.
PartField~\cite{11443382} and PA3FF~\cite{ICLR2026_52eebfd4} learn part-aware feature fields, the latter as the representation of a manipulation policy for articulated objects.
These methods identify which points belong to a part, whereas GRS trains point evidence through the object-relative position and direction that both tasks use.

\paragraph{3D language models and grounding.}
PointLLM~\cite{xu2024pointllm} and ShapeLLM~\cite{qi2024shapellm} describe objects and answer questions about them.
PartLLM~\cite{zhu2026partllmunifiedmultimodalfoundation} is a multimodal foundation model for 3D part segmentation.
Scene grounding methods localize referred objects~\cite{chen2020scanrefer,achlioptas2020referit3d}.
GAPartNet~\cite{10203924} estimates the poses of actionable parts, and SoFar~\cite{NEURIPS2025_6ea83df3} predicts the semantic orientation of an object, such as the direction a part faces.
Our target is instead where a part sits on the object.

\paragraph{Coordinate readouts.}
Integral regression~\cite{sun2018integral} and DSNT~\cite{nibali2018numerical} learn coordinates through expectations of spatial predictions, and DSNT also regularizes the heatmap's distribution.
GRS extends coordinate supervision with relative position error and explicit direction supervision on parts eligible for reorientation, aligning evidence learning with the geometric quantities used by both tasks.
Unlike direct offset regression or the per-point votes of VoteNet~\cite{qi2019deep}, our evidence weights the observed points themselves.
GRS acts on this evidence rather than on the encoder, so it applies to point architectures~\cite{8099499,NIPS2017_d8bf84be,NEURIPS2022_9318763d,10658198} and pretrained representations~\cite{9880161,11094293,zhang2026utonia} alike.

\paragraph{Rotation.}
Equivariant networks~\cite{Deng_2021_ICCV,thomas2018tensor} and learned canonicalization~\cite{kaba2023equivariance} build equivariance or invariance into the model.
Test time augmentation averages predictions over transformed inputs~\cite{9710313}, as GeoRead3D does.

\section{Method}
\label{sec:method}

\begin{figure*}[t]
  \centering
  \includegraphics[width=\textwidth]{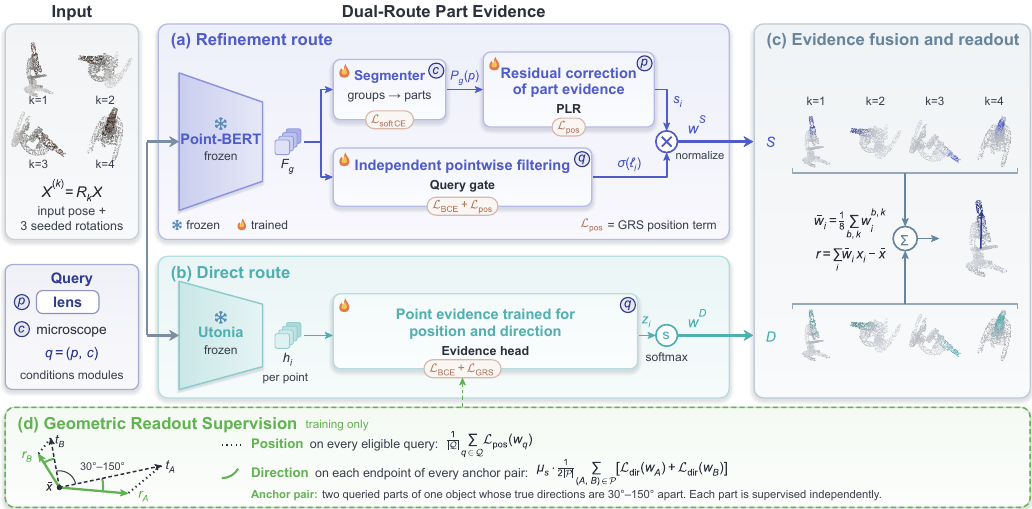}
  \caption{\textbf{Dual-Route Part Evidence with GRS.}
  (a)~The refinement route corrects a segmenter's group evidence with a Part Localization Refiner (PLR) and filters points with a query gate.
  (b)~The direct route predicts point evidence from frozen Utonia features.
  (c)~Evidence from both routes and four rotated copies of the input is averaged per point and read out once.
  (d)~During training, GRS supervises position on every queried part and direction on both parts of each anchor pair.
  Heatmaps show evidence for the lens of a held out 3DCoMPaT microscope.}
  \label{fig:method}
\end{figure*}

GeoRead3D predicts a query-conditioned distribution over the input points and reads out the part's location relative to the object center.
GRS applies position supervision to every queried part and direction supervision to the parts of anchor pairs.
Dual-Route Part Evidence produces the evidence along the two routes of \cref{fig:method}.

\subsection{Task and Evidence Representation}
\label{sec:setup}

\paragraph{Task.}
The input is a colored point cloud $X=\{x_i\}_{i=1}^{N}$, centered and scaled to unit maximum radius, together with the object category $c$ and a part description.
A fixed language model uses the category and candidate names, without seeing the point cloud, to map the description to a part name $p$ in the coarse 3DCoMPaT taxonomy~\cite{NEURIPS2024_f4f094d1}.
The output $r\in\mathbb{R}^3$ locates part $p$ relative to the object center $\bar x$, the centroid of all points.
Its target $t=\frac{1}{|\mathcal{I}_p|}\sum_{i\in \mathcal{I}_p}x_i-\bar x$ is the centroid of the annotated part points $\mathcal{I}_p$ relative to $\bar x$.
Pointing uses only the direction of $r$.

Two-anchor reorientation, or reorientation for short, receives descriptions of two parts $A$ and $B$ and commanded unit directions $u_A$ and $u_B$ in the input frame.
Hats denote unit vectors.
For evaluation, we generate jointly attainable commands $u_J=R^*\hat t_J$ for $J\in\{A,B\}$ from one target rotation $R^*$.
We estimate the object rotation by aligning the predicted unit directions $(\hat r_A,\hat r_B)$ with $(u_A,u_B)$ using Kabsch~\cite{kabsch1976solution}.

\paragraph{Point evidence and readout.}
Both routes express a part as point evidence, a distribution $w$ over the $N$ points with $w_i\ge0$ and $\sum_iw_i=1$, and a fixed readout turns it into a location,
\begin{equation}
  \Ro(w)=\sum_{i=1}^{N}w_i\,x_i-\bar x ,
  \label{eq:readout}
\end{equation}
the weighted centroid of the evidence relative to the object center~\cite{sun2018integral,nibali2018numerical}.
It is linear in $w$ and differentiable.

\subsection{Geometric Readout Supervision}
\label{sec:readout}

\paragraph{Position term.}
For evidence $w$ of part $p$ with target $t$, the position term is
\begin{equation}
  \Lgeo(w)=\frac{\lVert\Ro(w)-t\rVert_2^2}{\max(\lVert t\rVert_2^2,\epsilon)} .
  \label{eq:lgeo}
\end{equation}
Normalizing by $\lVert t\rVert^2$ makes the loss a relative error, so parts near and far from the object center contribute on the same scale.
It also bounds the pointing error.
Since $\angle(r,t)\le\arcsin(\lVert r-t\rVert/\lVert t\rVert)$ whenever $\lVert r-t\rVert<\lVert t\rVert$, any $\Lgeo<\sin^2 22.5^\circ\approx0.15$ guarantees successful pointing.
Parts whose offset is shorter than $0.05$ are excluded as queries in training and evaluation because their direction is unstable under small location errors, and we call the remaining parts eligible.
The excluded parts' labels still supervise the segmenter.
For eligible parts, $\epsilon=10^{-12}$ is inactive.

\paragraph{Length and direction.}
With $\rho=\lVert r\rVert/\lVert t\rVert$ and $\psi=\angle(r,t)$,
\begin{equation}
  \Lgeo=1+\rho^2-2\rho\cos\psi ,
  \label{eq:coupling}
\end{equation}
so the loss depends on the direction through $2\rho(1-\cos\psi)$, which shrinks with the readout length.
The direction term keeps this sensitivity independent of the length,
\begin{equation}
  \Ldir(w)=\frac{\lVert\hat r-\hat t\rVert_2^2}{4\sin^2(\beta/2)} ,
  \label{eq:ldir}
\end{equation}
with $\hat r=\Ro(w)/\lVert\Ro(w)\rVert$ for a nonzero readout.
Since $\lVert\hat r-\hat t\rVert=2\sin(\psi/2)$, the term is below one exactly when $\psi<\beta$.
We use the reorientation threshold $\beta=30^\circ$.

\paragraph{Objective.}
Reorientation commands name two parts of one object whose true directions are $30^\circ$ to $150^\circ$ apart.
In a training batch, two queried parts of the same shape that satisfy this condition form an anchor pair (\cref{fig:method}d), and $\mathcal{P}$ denotes these pairs.
For the eligible queries $\mathcal{Q}$ of a batch with evidence $w_q$, GRS is
\begin{equation}
  \begin{split}
    \Lgrs={}&\frac{1}{|\mathcal{Q}|}\sum_{q\in\mathcal{Q}}\Lgeo(w_q)\\
    &+\frac{\mu_s}{2|\mathcal{P}|}\sum_{(A,B)\in\mathcal{P}}\bigl(\Ldir(w_A)+\Ldir(w_B)\bigr) .
  \end{split}
  \label{eq:grs}
\end{equation}
The direction term averages over anchor pairs, so a part without a partner receives no direction supervision and a part in several pairs receives more.
Only the direction term is set to zero when a batch contains no anchor pair, and the remaining objectives are unchanged.
At initialization the evidence is close to uniform, so the readout is close to zero and its direction is numerically unstable.
The weight $\mu_s$ is zero before step $1{,}000$ of the $4{,}000$ training steps and rises linearly to $0.3$ at step $1{,}500$.

\subsection{Dual-Route Part Evidence}
\label{sec:dualroute}

The refinement route corrects the evidence of a part segmenter, while the direct route predicts evidence without segmentation (\cref{fig:method}a,b).
Both weight the same input points.

\paragraph{Refinement route.}
\label{sec:segbranch}
A frozen Point-BERT~\cite{9880161}, the colored encoder of PointLLM~\cite{xu2024pointllm}, divides the points into groups and produces a feature $F_g$ for each group $g$.
A part segmenter conditioned on the category predicts part probabilities $P_g(\cdot)$ from these features.
It is trained only with soft label cross entropy against each group's histogram of part labels.
The Part Localization Refiner (PLR) redistributes the evidence for the queried part between groups with a learned residual in the log domain,
\begin{align}
  a_g&=\operatorname*{softmax}_g\bigl(\log\max(P_g(p),\epsilon)+\delta_g\bigr),
  \label{eq:plr}\\
  s_i&=\textstyle\sum_g a_g\,\mathbf{1}[i\in\mathcal{N}_g]\,/\,|\mathcal{N}_g|,
  \nonumber
\end{align}
where $\epsilon=10^{-12}$ as in \cref{eq:lgeo}, $\mathcal{N}_g$ holds the points of group $g$, and points in no group receive $s_i=0$.
A small MLP predicts the residual $\delta_g$ from the group's features and probabilities and the queried part.
Its last layer is initialized to zero, so PLR starts exactly at the segmenter's own readout.
For PLR's training, two further segmenters are each trained on half of the training shapes.
PLR is trained with $\Lgeo(s)$ alone on each segmenter's probabilities for the half it did not see, so it learns to correct a segmenter's errors on unseen shapes.

PLR spreads each group's mass uniformly over its points, so it cannot distinguish points within a group.
An independent query gate therefore predicts a logit $\ell_i$ for every point from the group features, the point's offset from its group centers and its color.
It is conditioned on $q=(p,c)$ with FiLM~\cite{perez2018film} and attention across groups, and it does not see the segmenter or PLR.
The gate weights PLR's evidence,
\begin{equation}
  w^S_i=\frac{s_i\,\sigma(\ell_i)}{\sum_j s_j\,\sigma(\ell_j)} ,
  \label{eq:gate}
\end{equation}
where $\sigma$ is the logistic sigmoid.
The gate is trained through its own readout, with class balanced binary cross entropy (BCE) plus $\Lgeo(\mathrm{softmax}(\ell))$, not through \cref{eq:gate}.

\paragraph{Direct route.}
\label{sec:directbranch}
A frozen Utonia encoder~\cite{zhang2026utonia} provides a feature $h_i$ for each point (\cref{fig:method}b).
An evidence head conditioned on $q$ maps it to a logit $z_i$, giving $w^D=\mathrm{softmax}(z)$.
The head is trained with class balanced BCE on the part mask plus $\Lgrs+\Ldist$ on $w^D$.
It is the only module that receives the direction and distribution terms, while PLR and the gate receive the position term.
The auxiliary term $\Ldist$ penalizes the kernel-smoothed difference between the evidence and the uniform distribution on the part's points.

\subsection{Evidence Fusion and Readout}
\label{sec:averaging}

At inference, each route runs on four rotated copies of the input, $X^{(k)}=R_kX$, which we call poses: the input itself and three seeded uniformly random rotations~\cite{mezzadri2006generate}.
All poses contain the same points in the same order, so every route and pose yields a distribution over the same input points without correspondence estimation.
GeoRead3D averages the eight distributions with equal weights and reads out once (\cref{fig:method}c),
\begin{align}
  \bar w_i&=\frac18\sum_{b\in\{S,D\}}\sum_{k=1}^{4}w_i^{b,k},
  \label{eq:average}\\
  r=\Ro(\bar w)&=\frac18\sum_{b,k}R_k^{\top}\,\Ro^{(k)}\!\bigl(w^{b,k}\bigr),
  \label{eq:average_readout}
\end{align}
where $\Ro^{(k)}$ is the readout in pose $k$.
By the linearity of \cref{eq:readout}, averaging the evidence equals averaging the readouts rotated back to the input frame.
As $\Lgeo$ is convex in the readout, the fused position loss is at most the mean position loss of the eight outputs that enter the fusion.
For the refinement route, this output is the gated PLR evidence.
This bound concerns the position loss, and the ablations in \cref{sec:exp_ablation} measure the effect of aggregation on both tasks.

\subsection{Analysis of Two-Anchor Reorientation}
\label{sec:reorient_analysis}

Aligning both anchors within an angular tolerance does not necessarily place the whole object near its target orientation.
Let the true anchor directions $\hat t_A$ and $\hat t_B$ be $\theta$ apart, and let $\hat n$ be the unit normal of their plane, along $\hat t_A\times\hat t_B$.
Each predicted direction $\hat r_J$ deviates from $\hat t_J$ by an in-plane rotation $\gamma_J$, positive about $\hat n$, and a tilt $\zeta_J$ out of the plane, positive toward $\hat n$.
Let $e_A$ and $e_B$ be the angles between the true part directions and their commands after the estimated rotation, and $\phi$ the angle between the estimated rotation and $R^*$.
To first order in these deviations,
\begin{equation}
  \begin{aligned}
    e_{A,B}^2&=\bar\gamma^2+\zeta_{A,B}^2,\\
    \phi^2&=\bar\gamma^2+\frac{\zeta_+^2}{1+\cos\theta}+\frac{\zeta_-^2}{1-\cos\theta},
  \end{aligned}
  \label{eq:reorient}
\end{equation}
where $\bar\gamma=(\gamma_A+\gamma_B)/2$, and $\zeta_\pm=(\zeta_A\pm\zeta_B)/\sqrt2$ collect tilts to the same side ($+$) and to opposite sides ($-$).
Both criteria include the common in-plane rotation $\bar\gamma$ and ignore the relative in-plane error $\gamma_A-\gamma_B$.
They differ in the out-of-plane errors.
The anchor criterion does not amplify either anchor's tilt as $\theta$ changes.
The whole-object error instead amplifies $\zeta_-$ by $1/\sqrt{1-\cos\theta}$ when the anchors are close together, and $\zeta_+$ by $1/\sqrt{1+\cos\theta}$ when they point nearly opposite ways.
With the exact solver, true anchors $30^\circ$ apart with opposite tilts of $15^\circ$ leave each true anchor within approximately $11.6^\circ$ of its command, yet the whole-object rotation error is approximately $46^\circ$.
We therefore evaluate anchor alignment, which the commands specify, and whole-object rotation, which decides whether the object ends in its target pose.

\section{Experiments}
\label{sec:experiments}

\subsection{Experimental Setup}
\label{sec:exp_setup}

\paragraph{Data and queries.}
All trained modules and baselines use the $14{,}350$ training shapes of 3DCoMPaT~\cite{NEURIPS2024_f4f094d1} with its coarse taxonomy of $200$ categories and $118$ part names, and every object is represented by $2{,}048$ colored points.
Settings were chosen on held out 3DCoMPaT shapes and other PartNeXt shapes.
Each eligible part is queried with a synonym and a functional description, written once per category by a language model, and a shared mapper (Claude Haiku~4.5) gives every method the same part-name predictions, so the scores include mapping errors.
Our PartNeXt test split~\cite{NEURIPS2025_579636b6} holds $4{,}496$ chair and sofa shapes held out from development, with $26{,}782$ pointing queries and $17{,}382$ reorientation tasks.
Both categories are in the training taxonomy, and the mapper names the annotated part for all of its descriptions.
A sample of Artiverse~\cite{Iliash_2026_CVPR}, fixed before evaluation, gives $4{,}472$ pointing queries on $1{,}708$ shapes from $20$ categories and $602$ reorientation tasks on $301$ shapes.
Its source collections do not overlap our training data.

\paragraph{Protocol and metrics.}
Every test shape is shown in one seeded uniformly random rotation.
Pointing succeeds within $22.5^\circ$ of the true direction, and errors above $90^\circ$ count as severe.
Reorientation succeeds when, after the estimated rotation, both true part directions lie within $30^\circ$ of their commands.
Whole-object rotation succeeds when the geodesic angle between the estimated rotation and $R^*$ is at most $30^\circ$.
Undefined predictions count as failures.
Trained methods use three seeds, reported as mean and standard deviation.
Differences come with 95\% intervals from a paired bootstrap over shapes and seeds.
We call a difference significant when its interval excludes zero.

\begin{table*}[t]
  \centering
  \caption{\textbf{Comparison with prior methods on PartNeXt and Artiverse.}
  Success and severe error rates in percent on rotated input with the shared mapper and four poses, or one pose for COPS.
  Trained rows give the mean and standard deviation over three seeds, and bold marks the best mean.
  Params counts the networks run at inference, excluding cached text embeddings and the shared mapper.
  Time is the median latency per object on one GPU.
  Whole is whole-object rotation success within $30^\circ$.}
  \label{tab:comparison}
  \scriptsize
  \setlength{\tabcolsep}{4pt}
  \adjustbox{max width=\textwidth}{\begin{tabular}{@{}llcccccccccc@{}}
\toprule
 & & \multicolumn{4}{c}{PartNeXt test split} & \multicolumn{4}{c}{Artiverse} & & \\
\cmidrule(lr){3-6}\cmidrule(lr){7-10}
Method & Variant & Point. & Reorient. & Whole & Severe & Point. & Reorient. & Whole & Severe & Params & Time (ms) \\
\midrule
COPS~\cite{10943567} (WACV'25) & Released & 18.5 & 6.4 & 5.7 & 50.8 & 14.8 & 3.7 & 3.2 & 54.7 & 236.2M & 712 \\
Find3D~\cite{11444019} (ICCV'25) & Released & 38.4 & 26.0 & 23.0 & 23.0 & 26.5 & 13.3 & 11.1 & 33.7 & 46.2M & 282 \\
CoSMo3D~\cite{jin2026cosmo3d} (CVPR'26) & Released & 41.8 & 27.7 & 23.6 & 19.6 & 37.2 & 16.6 & 14.1 & 26.3 & 57.5M & 251 \\
PatchAlign3D~\cite{Hadgi_2026_CVPR} (CVPR'26) & Released & 45.5 & 33.4 & 29.7 & 19.3 & 29.7 & 18.3 & 16.4 & 37.6 & 22.4M & 57 \\
\midrule
Point-BERT~\cite{9880161} (CVPR'22) & Frozen + mask head & 64.7\std{1.1} & 53.8\std{1.8} & 48.2\std{1.8} & 5.8\std{0.5} & 52.5\std{0.9} & 27.1\std{0.6} & 24.9\std{0.4} & 16.5\std{0.1} & 22.1M & 140 \\
Find3D~\cite{11444019} (ICCV'25) & Frozen + mask head & 39.2\std{0.8} & 26.8\std{0.8} & 23.8\std{0.7} & 23.5\std{1.0} & 33.2\std{0.3} & 7.6\std{1.0} & 6.5\std{1.3} & 30.9\std{0.4} & 46.7M & 252 \\
PatchAlign3D~\cite{Hadgi_2026_CVPR} (CVPR'26) & Frozen + mask head & 51.5\std{0.9} & 37.1\std{1.4} & 31.7\std{1.8} & 9.6\std{0.3} & 37.9\std{0.4} & 17.8\std{1.7} & 16.6\std{1.4} & 30.9\std{0.3} & 23.1M & 46 \\
Utonia~\cite{zhang2026utonia} (ICML'26) & Frozen + mask head & 68.4\std{1.0} & 60.8\std{1.6} & 54.0\std{1.3} & 4.2\std{0.2} & 51.0\std{0.8} & 25.0\std{1.4} & 23.2\std{1.1} & 18.4\std{0.6} & 138.0M & 147 \\
\midrule
Point-BERT~\cite{9880161} (CVPR'22) & Fine-tuned & 67.8\std{0.3} & 59.5\std{1.0} & 54.2\std{0.9} & 6.7\std{0.2} & 52.7\std{0.4} & 24.4\std{0.8} & 22.8\std{0.5} & 17.9\std{0.7} & 22.1M & 41 \\
PointNeXt-S~\cite{NEURIPS2022_9318763d} (NeurIPS'22) & From scratch & 43.3\std{1.0} & 30.8\std{0.8} & 27.4\std{0.9} & 16.8\std{0.6} & 37.1\std{1.0} & 16.3\std{0.7} & 15.3\std{1.2} & 30.7\std{0.9} & 22.6M & 52 \\
PointLLM-7B~\cite{xu2024pointllm} (ECCV'24) & Fine-tuned & 53.6\std{1.2} & 40.8\std{1.5} & 36.6\std{1.5} & 10.2\std{0.8} & 49.1\std{1.0} & 20.7\std{1.4} & 18.1\std{0.7} & 18.0\std{1.3} & 6.78B & 483 \\
Utonia~\cite{zhang2026utonia} (ICML'26) & Fine-tuned & 63.6\std{1.7} & 55.8\std{2.6} & 50.6\std{2.4} & 7.6\std{0.6} & 44.1\std{1.4} & 21.3\std{1.7} & 19.8\std{1.8} & 23.8\std{0.8} & 137.6M & 159 \\
\midrule
\textbf{GeoRead3D} (ours) & --- & \textbf{74.1\std{0.4}} & \textbf{71.2\std{0.3}} & \textbf{65.3\std{0.4}} & \textbf{3.0\std{0.1}} & \textbf{60.0\std{0.8}} & \textbf{34.9\std{2.7}} & \textbf{32.2\std{2.3}} & \textbf{12.7\std{0.8}} & 160.6M & 307 \\
\bottomrule
\end{tabular}
}
\end{table*}

\begin{table}[t]
  \centering
  \caption{\textbf{Comparison with prior methods across angular thresholds.}
  Success rates in percent, with each threshold applied to pointing, reorientation and whole-object rotation.
  Values are means under the settings of \cref{tab:comparison}, and bold marks the best mean.
  Each baseline entry gives the highest mean among compared methods: frozen Point-BERT (P) or Utonia (U) with a mask head, or fine-tuned Point-BERT (F).}
  \label{tab:thresholds}
  \scriptsize
  \setlength{\tabcolsep}{2pt}
  \begin{tabular*}{\columnwidth}{@{\extracolsep{\fill}}llcccccc@{}}
\toprule
 & & \multicolumn{3}{c}{PartNeXt test split} & \multicolumn{3}{c}{Artiverse} \\
\cmidrule(lr){3-5}\cmidrule(l){6-8}
 & & Point. & Reorient. & Whole & Point. & Reorient. & Whole \\
\midrule
$15^\circ$ & GeoRead3D & \textbf{61.8} & \textbf{44.9} & \textbf{39.7} & \textbf{52.2} & \textbf{20.9} & \textbf{19.4} \\
 & Strongest baseline & 57.3\textsuperscript{F} & 38.2\textsuperscript{F} & 34.0\textsuperscript{F} & 45.5\textsuperscript{F} & 16.7\textsuperscript{U} & 15.7\textsuperscript{U} \\
\midrule
$22.5^\circ$ & GeoRead3D & \textbf{74.1} & \textbf{60.4} & \textbf{54.3} & \textbf{60.0} & \textbf{27.8} & \textbf{25.4} \\
 & Strongest baseline & 68.4\textsuperscript{U} & 50.3\textsuperscript{U} & 45.1\textsuperscript{F} & 52.7\textsuperscript{F} & 20.7\textsuperscript{U} & 19.5\textsuperscript{P} \\
\midrule
$30^\circ$ & GeoRead3D & \textbf{82.0} & \textbf{71.2} & \textbf{65.3} & \textbf{65.9} & \textbf{34.9} & \textbf{32.2} \\
 & Strongest baseline & 76.6\textsuperscript{U} & 60.8\textsuperscript{U} & 54.2\textsuperscript{F} & 58.2\textsuperscript{F} & 27.1\textsuperscript{P} & 24.9\textsuperscript{P} \\
\midrule
$45^\circ$ & GeoRead3D & \textbf{90.2} & \textbf{83.2} & \textbf{77.3} & \textbf{74.7} & \textbf{45.3} & \textbf{42.0} \\
 & Strongest baseline & 86.2\textsuperscript{U} & 75.5\textsuperscript{U} & 67.9\textsuperscript{U} & 66.9\textsuperscript{P} & 38.0\textsuperscript{P} & 34.3\textsuperscript{P} \\
\bottomrule
\end{tabular*}

\end{table}

\paragraph{Baselines.}
The released part segmenters Find3D~\cite{11444019}, PatchAlign3D~\cite{Hadgi_2026_CVPR}, CoSMo3D~\cite{jin2026cosmo3d} and COPS~\cite{10943567} are used unchanged.
Their location is the centroid of their point scores for the queried name, sharpened by a power tuned on training shapes, or of the mask for COPS.
Find3D and CoSMo3D receive their native mesh samples, and CoSMo3D was trained on our training shapes with its own labels.
The trained baselines use our data and labels: mask heads on four frozen encoders; fine-tuned Point-BERT and Utonia; PointNeXt-S~\cite{NEURIPS2022_9318763d} trained from scratch with its official recipe on 544K samples; and PointLLM-7B~\cite{xu2024pointllm}.
All trained baselines other than PointNeXt-S use at most 128K training samples.
PointLLM has no grounding output, so it receives a LoRA-trained readout of its attention to the point groups.
To check sensitivity to rotation augmentation, we retrain fine-tuned Point-BERT, fine-tuned Utonia and PointNeXt-S with fresh instead of fixed random rotations, and none improves.

\subsection{Comparison with Prior Methods}
\label{sec:exp_comparison}

\paragraph{Main comparison.}
\Cref{tab:comparison} compares all methods at the success thresholds of the two tasks, $22.5^\circ$ for pointing and $30^\circ$ for reorientation and whole-object rotation.
GeoRead3D has the highest mean success rates and the lowest mean severe error rates on both datasets.
It leads the strongest baseline by $5.8$ pointing, $10.4$ reorientation and $11.1$ whole-object points on PartNeXt, and by $7.3$, $7.8$ and $7.3$ points on Artiverse.
These strongest baselines, frozen encoders with a mask head and fine-tuned Point-BERT, train on the same data and labels as GeoRead3D, so the margins do not come from the training data.
We test category-mean directions and five-shape retrieval with true poses and correct part names.
Category-mean directions reach $55.5\%$ pointing on PartNeXt and $30.7\%$ on Artiverse; retrieval reaches $62.0\%$ and $44.7\%$, respectively.
Both remain below GeoRead3D.
PartLLM~\cite{zhu2026partllmunifiedmultimodalfoundation} may have seen Artiverse during training, so we compare it on a subset of our PartNeXt test split that excludes its training objects.
On that subset, with correct part names, GeoRead3D reaches $75.1\%$ pointing and $73.5\%$ reorientation against $71.9\%$ and $65.8\%$.
Its reorientation lead of $7.7$ points is significant, and its severe error rate is $2.4\%$, compared with $9.8\%$ for PartLLM.

\paragraph{Shared angular thresholds.}
To test whether the advantage depends on these thresholds, \cref{tab:thresholds} scores every metric at one shared threshold of $15^\circ$, $22.5^\circ$, $30^\circ$ or $45^\circ$ and compares GeoRead3D with the strongest compared method in each cell.
GeoRead3D leads in all $24$ cells, by $4.0$ to $11.1$ points on PartNeXt and $3.7$ to $7.8$ points on Artiverse.
The strongest compared method changes with the threshold, metric and dataset, so the advantage depends neither on the chosen tolerances nor on a single comparator.

GeoRead3D trains part evidence for the locations and directions used by both tasks and combines this evidence across routes and poses, reducing reliance on any single prediction.
The following controlled comparisons examine how these design choices contribute to its advantage.

\subsection{Evaluating GRS}
\label{sec:exp_objectives}

\begin{table}[t]
  \centering
  \caption{\textbf{Evaluation of GRS across encoders, objectives and input poses.}
  Success rates in percent on rotated input with correct part names, reported as mean and standard deviation over three seeds.
  Panels (a) and (b) use the PartNeXt test split with four poses; (c) uses both datasets with one input pose.
  In (a), Mask uses mask supervision alone, and + Position adds the position loss.
  In (b), frozen-encoder heads receive BCE and the listed objectives.
  Differences are relative to Mask in (a, c) and Full GRS in (b), in bold when significant.}
  \label{tab:objectives}\label{tab:readout}
  \scriptsize
  \setlength{\tabcolsep}{2pt}
  (a) Position term across encoders and training settings\\[2pt]
  \begin{tabular*}{\columnwidth}{@{\extracolsep{\fill}}rlcc@{\hspace{3pt}}c@{\hspace{3pt}}c@{}}
\toprule
 & & \multicolumn{2}{c}{Pointing} & \multicolumn{2}{c}{$\Delta$} \\
\cmidrule(lr){3-4}\cmidrule(l){5-6}
\# & Encoder & Mask & + Position & Point. & Reorient. \\
\midrule
\multicolumn{6}{@{}l}{Frozen + head} \\
1 & Point-BERT~\cite{9880161} (CVPR'22) & 64.7\std{1.1} & 69.1\std{0.6} & \textbf{+4.4}\std{1.1} & \textbf{+8.9}\std{1.8} \\
2 & Find3D~\cite{11444019} (ICCV'25) feat. & 39.2\std{0.8} & 46.5\std{2.5} & \textbf{+7.3}\std{2.4} & \textbf{+8.5}\std{2.0} \\
3 & PatchAlign3D~\cite{Hadgi_2026_CVPR} (CVPR'26) & 51.5\std{0.9} & 56.2\std{0.4} & \textbf{+4.7}\std{1.3} & \textbf{+6.0}\std{1.8} \\
4 & Utonia~\cite{zhang2026utonia} (ICML'26) & 68.4\std{1.0} & 70.3\std{0.9} & \textbf{+1.9}\std{1.0} & \textbf{+4.8}\std{2.2} \\
\midrule
\multicolumn{6}{@{}l}{Fine-tuned} \\
5 & Point-BERT~\cite{9880161} (CVPR'22) & 67.8\std{0.3} & 68.8\std{0.5} & \textbf{+1.0}\std{0.5} & \textbf{+2.3}\std{0.7} \\
6 & Utonia~\cite{zhang2026utonia} (ICML'26) & 63.6\std{1.7} & 65.3\std{1.8} & +1.7\std{1.5} & +3.0\std{3.1} \\
\midrule
\multicolumn{6}{@{}l}{From scratch} \\
7 & PointNeXt-S~\cite{NEURIPS2022_9318763d} (NeurIPS'22) & 28.0\std{2.0} & 31.6\std{2.2} & \textbf{+3.6}\std{2.6} & \textbf{+3.4}\std{3.3} \\
\bottomrule
\end{tabular*}
\\[6pt]
  (b) Training objectives on frozen encoders\\[2pt]
  \begin{tabular*}{\columnwidth}{@{\extracolsep{\fill}}>{\raggedright\arraybackslash}p{98pt}cccc@{}}
\toprule
 & & & \multicolumn{2}{c}{$\Delta$} \\
\cmidrule(l){4-5}
Objective & Point. & Reorient. & Point. & Reorient. \\
\midrule
\multicolumn{5}{@{}l}{Utonia~\cite{zhang2026utonia} (ICML'26), frozen + head} \\
DSNT coordinate loss~\cite{nibali2018numerical} & 69.6\std{0.5} & 63.9\std{0.9} & \textbf{$-$3.0}\std{0.4} & \textbf{$-$5.1}\std{1.3} \\
DSNT, matched weight & 71.9\std{0.2} & 65.9\std{0.3} & \textbf{$-$0.6}\std{0.2} & \textbf{$-$3.0}\std{0.5} \\
DSNT, $2\times$ matched weight & 71.2\std{0.3} & 65.1\std{0.6} & \textbf{$-$1.4}\std{0.6} & \textbf{$-$3.8}\std{1.1} \\
Position loss & 70.3\std{0.9} & 65.6\std{1.0} & \textbf{$-$2.3}\std{0.6} & \textbf{$-$3.3}\std{0.8} \\
+ Direction, every part & 71.2\std{0.7} & 67.2\std{0.5} & \textbf{$-$1.3}\std{0.5} & \textbf{$-$1.8}\std{0.4} \\
+ Direction, parts with a partner & 72.1\std{0.5} & 68.9\std{0.3} & \textbf{$-$0.5}\std{0.3} & 0.0\std{0.2} \\
Full GRS & 72.6\std{0.2} & 69.0\std{0.5} & --- & --- \\
+ Rotation-level pair loss & 72.5\std{0.2} & 69.4\std{0.2} & $-$0.1\std{0.1} & +0.4\std{0.7} \\
\midrule
\multicolumn{5}{@{}l}{Point-BERT~\cite{9880161} (CVPR'22), frozen + head} \\
DSNT coordinate loss~\cite{nibali2018numerical} & 67.1\std{0.3} & 57.4\std{0.6} & \textbf{$-$4.5}\std{0.1} & \textbf{$-$9.1}\std{0.6} \\
DSNT, matched weight & 70.0\std{0.4} & 60.9\std{0.8} & \textbf{$-$1.6}\std{0.5} & \textbf{$-$5.5}\std{1.4} \\
DSNT, $2\times$ matched weight & 70.1\std{0.6} & 61.3\std{0.5} & \textbf{$-$1.5}\std{0.2} & \textbf{$-$5.1}\std{0.8} \\
Position loss & 69.1\std{0.6} & 62.7\std{1.0} & \textbf{$-$2.4}\std{0.5} & \textbf{$-$3.7}\std{0.9} \\
+ Direction, every part & 70.1\std{0.9} & 63.8\std{1.6} & \textbf{$-$1.4}\std{0.5} & \textbf{$-$2.6}\std{0.6} \\
+ Direction, parts with a partner & 71.3\std{0.5} & 65.7\std{1.5} & $-$0.3\std{0.2} & \textbf{$-$0.7}\std{0.6} \\
Full GRS & 71.5\std{0.4} & 66.4\std{0.9} & --- & --- \\
+ Rotation-level pair loss & 71.5\std{0.3} & 67.3\std{0.8} & 0.0\std{0.4} & \textbf{+0.8}\std{0.4} \\
\bottomrule
\end{tabular*}
\\[6pt]
  (c) One frozen encoder and one input pose\\[2pt]
  \begin{tabular*}{\columnwidth}{@{\extracolsep{\fill}}lcccc@{}}
\toprule
 & \multicolumn{2}{c}{PartNeXt test split} & \multicolumn{2}{c}{Artiverse} \\
\cmidrule(lr){2-3}\cmidrule(l){4-5}
Objective & Point. & Reorient. & Point. & Reorient. \\
\midrule
\multicolumn{5}{@{}l}{Utonia~\cite{zhang2026utonia} (ICML'26), frozen + head} \\
Mask & 62.7\std{0.8} & 52.1\std{0.7} & 47.9\std{0.4} & 20.8\std{1.0} \\
Full GRS & 67.8\std{0.4} & 61.2\std{0.8} & 54.7\std{0.5} & 25.6\std{0.6} \\
$\Delta$ & \textbf{+5.1}\std{0.9} & \textbf{+9.2}\std{1.5} & \textbf{+6.9}\std{0.9} & \textbf{+4.8}\std{0.8} \\
\midrule
\multicolumn{5}{@{}l}{Point-BERT~\cite{9880161} (CVPR'22), frozen + head} \\
Mask & 62.3\std{1.1} & 49.9\std{1.4} & 50.4\std{0.8} & 23.6\std{0.9} \\
Full GRS & 69.3\std{0.4} & 62.6\std{0.9} & 56.0\std{1.5} & 31.7\std{5.3} \\
$\Delta$ & \textbf{+7.0}\std{0.9} & \textbf{+12.7}\std{0.5} & \textbf{+5.6}\std{2.2} & \textbf{+8.1}\std{5.6} \\
\bottomrule
\end{tabular*}

\end{table}

We evaluate GRS on individual encoders to separate the effect of supervision from the dual-route system (\cref{tab:objectives}).
Correct part names isolate localization, and each comparison matches the encoder, training data, number of training samples ($128$K) and seeds.

\paragraph{Position supervision.}
The position term improves both tasks on all four frozen encoders (\cref{tab:objectives}a), even though mean mask IoU decreases, because it supervises the centroid that the tasks use.
In three of these four comparisons, the evidence mass on the part's points increases and the centroid error decreases.
Its benefit also extends to fine-tuned Point-BERT and PointNeXt-S trained from scratch.

\paragraph{Direction supervision.}
Unlike the position term, the direction term's penalty does not shrink with the predicted offset length (\cref{eq:coupling,eq:ldir}), and it improves pointing on both encoders even when applied to every part (\cref{tab:objectives}b).
Full GRS applies it to the parts of anchor pairs, each supervised independently and weighted by its number of pairs.
Most of its gain over every-part supervision comes from selecting parts with a valid reorientation partner: with the same direction-loss weight and schedule, supervising these parts with equal weights significantly improves both tasks over every-part supervision on both encoders.
Weighting by pair count adds smaller and less consistent gains.

\paragraph{Alternative objectives.}
Full GRS outperforms DSNT's Euclidean coordinate loss in both tasks on both encoders, with the head and training setup fixed (\cref{tab:objectives}b).
This holds at DSNT's own weight, at the weight matched to the position loss's initial gradient ($16.3$ for Utonia and $16.2$ for Point-BERT), and at twice that weight.
The rotation-level pair loss, $\sin^2(\phi/2)$ per anchor pair instead of full GRS's per-part direction terms, adds a small reorientation gain on PartNeXt with the same pairs, weight and schedule, significant for Point-BERT, and no significant pointing gain (\cref{tab:objectives}b).
This supports per-part direction supervision as the simpler objective.

\paragraph{Readout rules.}
Each head reads out with a rule selected on held out training shapes, and reading all heads through a common thresholded mask on the input pose removes the position term's pointing gain on PartNeXt.
Under this readout, direction supervision still improves both tasks over the position term, so it does not rely on soft weighting.

\paragraph{Evidence and direct regression.}
Matched regression heads share the evidence heads' features, data, seeds and budget, predict the mask from the same trunk, and receive the same BCE, position and paired direction supervision.
With direction supervision and four poses, the evidence readout points more accurately by $1.3$ to $2.6$ points across both encoders on PartNeXt and Artiverse, significantly in three of the four comparisons.
Reorientation shows no consistent advantage for either readout.

With one frozen encoder and a single input pose, GRS improves pointing by $5.1$--$7.0$ points and reorientation by $4.8$--$12.7$ points on PartNeXt and Artiverse across both encoders, without a second evidence route or aggregation over poses (\cref{tab:objectives}c).

\subsection{Geometry of Part Localization}
\label{sec:exp_geometry}

\begin{figure}[t]
  \centering
  \includegraphics[width=\columnwidth]{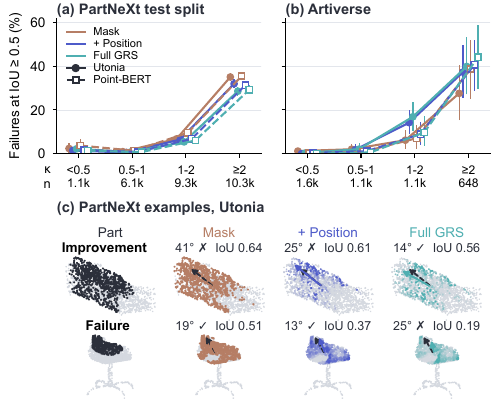}
  \caption{\textbf{Mask overlap and pointing by part geometry.}
  (a, b)~Pointing failures among masks with IoU $\ge0.5$, grouped by $\kappa$, the part's spread divided by its distance from the object center, for heads on frozen encoders.
  Input pose and correct part names, with 95\% intervals and $n$ queries per group.
  (c)~Two examples chosen by a fixed rule, true directions dashed and readouts solid.}
  \label{fig:kappa}
\end{figure}

\Cref{fig:kappa} examines when mask overlap predicts pointing accuracy.
We group parts by $\kappa=d_p/\lVert t\rVert$.
Here $d_p$ is the root mean square distance of the part's points from their centroid, and $\lVert t\rVert$ is the centroid's distance from the object center.
The label $n$ gives the total number of pointing queries in each group.
Below $\kappa=1$, nearly every mask with IoU of at least $0.5$ points correctly.
From $\kappa=2$ upward, a quarter to a half of them fail, for every head and both encoders on both datasets.
High mask overlap does not ensure an accurate direction when part spread is large relative to its offset from the object center.
On PartNeXt, paired direction supervision raises pointing over the position term most at $\kappa\ge2$, by $3.8$ and $3.3$ points for Utonia and Point-BERT over all queries in this group.
The corresponding pointing difference on Artiverse is not significant.

\Cref{fig:kappa}c shows how GRS changes the evidence.
In the improvement example, paired direction supervision shifts weight toward the far end of a part the mask head already covers well, so the readout turns toward the true direction despite lower IoU.
In the failure example, the GRS evidence spreads onto a neighboring part and pulls the readout away from the target.

\subsection{System Comparisons}
\label{sec:exp_ablation}

\begin{table}[t]
  \centering
  \caption{\textbf{Comparison of GeoRead3D variants on the PartNeXt test split.}
  Success and severe error rates in percent on rotated input with the shared mapper and four poses unless stated.
  Values are the mean and standard deviation over three seeds of the gate and evidence head, and bold marks the best mean.
  Whole is whole-object rotation success within $30^\circ$.}
  \label{tab:ablation}
  \scriptsize
  \setlength{\tabcolsep}{2pt}
  \begin{tabular}{@{}r>{\raggedright\arraybackslash}p{89pt}cccc@{}}
\toprule
\# & Variant & Point. & Reorient. & Whole & Severe \\
\midrule
1 & \textbf{GeoRead3D} & \textbf{74.1\std{0.4}} & \textbf{71.2\std{0.3}} & \textbf{65.3\std{0.4}} & 3.0\std{0.1} \\
\midrule
2 & Evidence head with the position term only & 73.1\std{0.2} & 69.8\std{0.4} & 63.8\std{0.2} & 3.1\std{0.1} \\
3 & Mask objectives in every module & 72.1\std{0.5} & 66.7\std{0.9} & 60.5\std{0.9} & 3.5\std{0.2} \\
\midrule
4 & Refinement route only & 70.2\std{0.2} & 63.8\std{0.5} & 57.3\std{0.4} & 4.8\std{0.3} \\
5 & Direct route only & 72.5\std{1.0} & 69.0\std{0.8} & 62.1\std{0.8} & \textbf{2.8\std{0.1}} \\
\midrule
6 & PLR replaced by sharpening & 72.7\std{0.5} & 68.5\std{0.2} & 62.2\std{0.3} & 3.2\std{0.1} \\
7 & No query gate & 72.4\std{0.6} & 69.5\std{0.2} & 63.1\std{0.3} & 3.4\std{0.1} \\
8 & Input pose only & 71.4\std{0.5} & 66.9\std{0.1} & 59.9\std{0.1} & 3.4\std{0.2} \\
\bottomrule
\end{tabular}

\end{table}

\paragraph{Supervision at fixed architecture.}
With the architecture unchanged, GeoRead3D's training objectives improve all three success rates on PartNeXt over mask objectives alone (\cref{tab:ablation}).
The paired direction and distribution terms of the evidence head add gains over its position term alone.

\paragraph{Combining evidence routes.}
Combining the trained routes improves all three success rates on PartNeXt over either route alone.
On Artiverse, the gains over the direct route are larger, at $3.6$ pointing and $6.6$ reorientation points, both significant, while the reorientation gain over the refinement route is not significant.
Fusion requires a second encoder and refinement modules; on PartNeXt, the direct route alone has a lower mean severe error rate.

\paragraph{Refinement and aggregation.}
Replacing PLR's learned redistribution with fixed sharpening of the same probabilities lowers all three success rates.
Removing the query gate lowers them as well, supporting pointwise filtering within groups.
Reading out only the input pose lowers all three, so aggregating across poses improves both tasks.

\subsection{Human-Written Descriptions}
\label{sec:exp_human}

\begin{table}[t]
  \centering
  \caption{\textbf{Comparison of model-written and human-written descriptions.}
  Success rates in percent under the settings of \cref{tab:comparison}, averaged over three writers.
  Both query sources are evaluated on identical tasks for each writer.
  Standard deviations are over three model seeds.
  Differences are human-written minus model-written, in bold when significant.}
  \label{tab:human}
  \scriptsize
  \setlength{\tabcolsep}{3pt}
  \adjustbox{max width=\columnwidth}{\begin{tabular}{@{}lrrrr@{}}
\toprule
Query source & \begin{tabular}[b]{@{}r@{}}PartNeXt\\pointing\end{tabular} & \begin{tabular}[b]{@{}r@{}}PartNeXt\\reorientation\end{tabular} & \begin{tabular}[b]{@{}r@{}}Artiverse\\pointing\end{tabular} & \begin{tabular}[b]{@{}r@{}}Artiverse\\reorientation\end{tabular} \\
\midrule
Model-written & 72.5\std{0.4} & 70.1\std{0.2} & 58.3\std{0.8} & 30.8\std{2.9} \\
Human-written, 3 writers & 69.9\std{0.4} & 64.9\std{0.2} & 59.0\std{1.0} & 32.7\std{3.5} \\
$\Delta$ & \textbf{$-$2.6} & \textbf{$-$5.2} & \textbf{+0.7} & +2.0 \\
\bottomrule
\end{tabular}
}
\end{table}

Model-written queries may underrepresent how users describe parts.
We evaluate queries from three native English speakers using the same mapper and GeoRead3D model on identical tasks for each writer (\cref{tab:human}).
Success decreases on PartNeXt but changes less on Artiverse.
The PartNeXt drop comes from two sofa descriptions mapped to unintended parts and reused across objects.
Replacing Haiku~4.5 with Haiku~5.5 or Sonnet~5.5 reduces the PartNeXt drops to $1.6$ pointing and $2.8$ reorientation points.

\section{Limitations}
\label{sec:limitations}

GeoRead3D grounds descriptions to parts of a fixed taxonomy with a known category.
It inherits the errors of its mapper, and a part name may denote several instances, which are located by the centroid of all their points.
Its inputs are complete point clouds sampled from meshes, and partial views, sensor noise and real scans remain untested.
Both tasks are evaluated offline without physical execution.
The position term's untuned weight may need adjustment for other mask losses.

\section{Conclusion}
\label{sec:conclusion}

Part segmentation quality alone does not determine how accurately a part is located.
Our analysis shows how part geometry affects the relationship between mask overlap and pointing accuracy.
Our two-anchor analysis also shows why successful anchor alignment can leave substantial whole-object rotation error.
GRS learns point evidence through its object-relative position and direction, with directional supervision focused on parts that can form valid reorientation commands.
Controlled experiments demonstrate gains with a single frozen encoder and a single input pose.
GeoRead3D combines evidence across routes and poses and achieves the highest pointing and reorientation success among the evaluated methods on both test sets.

\section*{LLM Usage Statement}
Language models are part of the evaluated pipeline.
Claude Opus~5.5 wrote the synonym and functional description of each part, Claude Sonnet~5 checked them, and Claude Haiku~4.5 is the mapper that turns each description into a part name.
The supplementary mapper comparison also uses Claude Haiku~5.5 and Claude Sonnet~5.5.
PointLLM and PartLLM are compared methods.
We also used Claude Code to write and debug experiment, figure and table code and to run and analyze experiments.
We used Claude Code and Codex with GPT-6-Astra (high reasoning effort) to refine the manuscript and to suggest revisions.
The authors directed this work, checked the code, results and text, and take full responsibility for the content.

{
    \small
    \bibliographystyle{ieeenat_fullname}
    \bibliography{main}
}

\end{document}